\documentclass[10pt,twocolumn,letterpaper]{article}

\usepackage{iccv}
\usepackage{times}
\usepackage{epsfig}
\usepackage{graphicx}
\usepackage{amsmath}
\usepackage{amssymb}
\usepackage{array,booktabs}
\usepackage{subfiles}
\usepackage[misc]{ifsym} 

\usepackage{threeparttable}
\usepackage[pagebackref=true,breaklinks=true,letterpaper=true,colorlinks,bookmarks=false]{hyperref}
\usepackage{multirow}
\usepackage{lscape}
\newcolumntype{M}[1]{>{\centering\arraybackslash}m{#1}}
\newcolumntype{N}{@{}m{0pt}@{}}
\usepackage{hyperref}

\newcounter{alphasect}
\def\alphainsection{0}

\let\oldsection=\section
\def\section{%
  \ifnum\alphainsection=1%
    \addtocounter{alphasect}{1}
  \fi%
\oldsection}%

\renewcommand\thesection{%
  \ifnum\alphainsection=1%
    \Alph{alphasect}
  \else%
    \arabic{section}
  \fi%
}%

\iccvfinalcopy 

\def\iccvPaperID{407} 

\ificcvfinal\fi
\begin{document}

\title{StructureFlow: Image Inpainting via Structure-aware Appearance Flow}

\author{Yurui Ren$^{1,2}$~~~Xiaoming Yu$^{1,2}$~~~Ruonan Zhang$^{2}$~~~Thomas H. Li$^{3,1}$~~~Shan Liu$^{4}$~~~{Ge Li \footnotesize{\Letter}}$^{1,2}$\\
$^1$School of Electronics and Computer Engineering, Peking University~~~$^2$Peng Cheng Laboratory~~~\\
$^3$Advanced Institute of Information Technology, Peking University~~~$^4$Tencent America~~~~~~~~~\\
{\tt\small yrren@pku.edu.cn~~~~xiaomingyu@pku.edu.cn~~~~zhangrn@pcl.ac.cn~~~}\\  {\tt\small tli@aiit.org.cn~~~~~shanl@tencent.com~~~~~~~~geli@ece.pku.edu.cn}
}

\maketitle

\begin{abstract}

Image inpainting techniques have shown significant improvements by using deep neural networks recently. However, most of them may either fail to reconstruct reasonable structures or restore fine-grained textures. In order to solve this problem, in this paper, we propose a two-stage model which splits the inpainting task into two parts: structure reconstruction and texture generation. In the first stage, edge-preserved smooth images are employed to train a structure reconstructor which completes the missing structures of the inputs. In the second stage, based on the reconstructed structures, a texture generator using appearance flow is designed to yield image details. Experiments on multiple publicly available datasets show the superior performance of the proposed network.
\end{abstract}

\section{Introduction}
\begin{figure}[t]
\begin{center}
\includegraphics[width=1\linewidth]{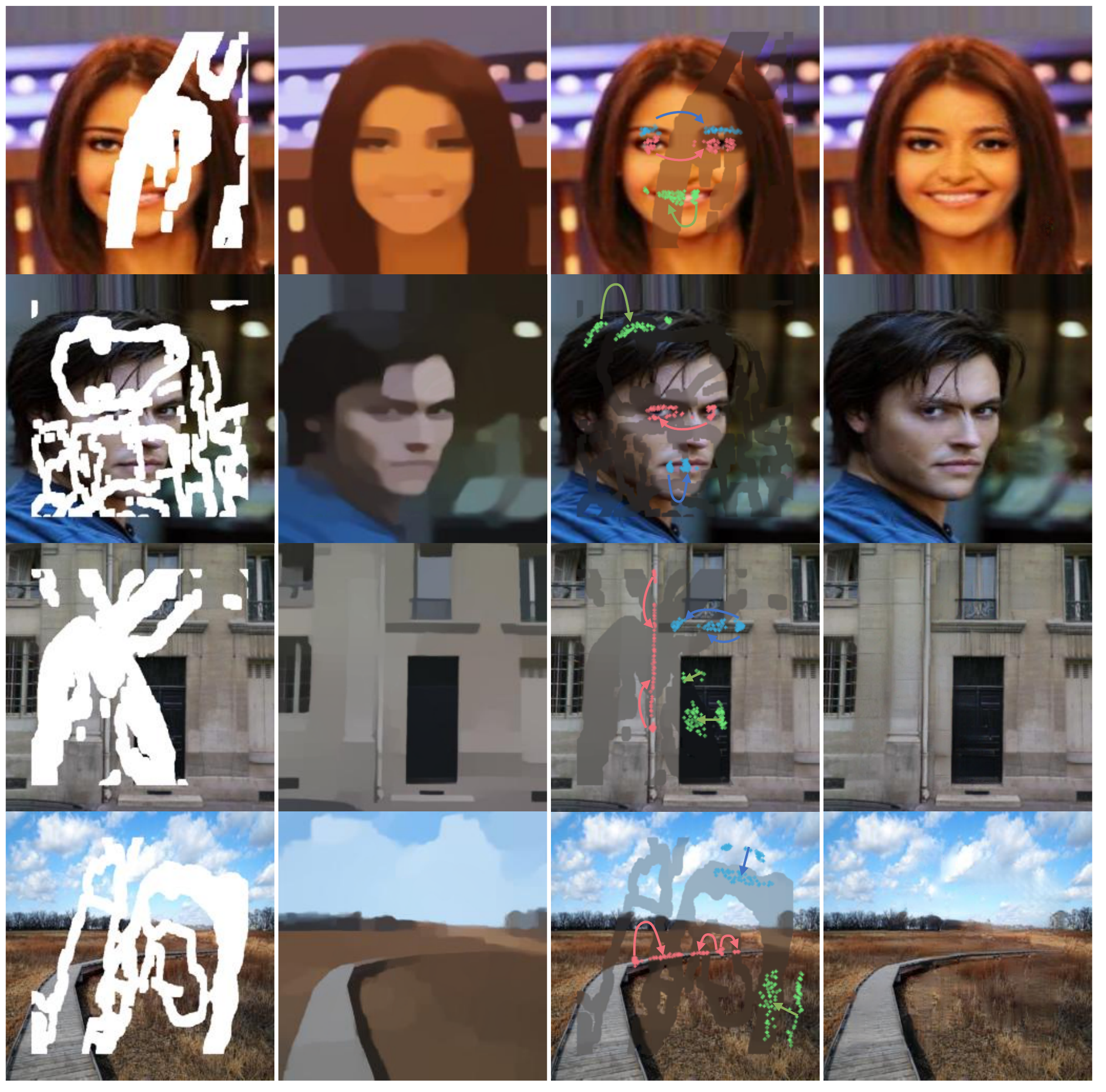}
\end{center}
   \caption{(From left to right) Input corrupted images, reconstructed structure images, visualizations of the appearance flow fields, final output images. 
   Our method first recovers global structures for missing regions, then generate textures by sampling features from existing regions according to the recovered structures. 
   To visualize the appearance flow fields, we plot the sample points of some typical missing regions. The arrows show the direction of the appearance flow.}
\label{fig:Formework}
\end{figure}

Image inpainting refers to generating alternative structures and textures for missing regions of corrupted input images and obtaining visually realistic results. It has a wide range of applications. For example, users can remove unwanted objects or edit contents of images by using inpainting techniques. A major challenge of image inpainting tasks is to generate correct structures and realistic textures. Some early patch-based works attempt to fill missing holes with image patches from existing regions~\cite{barnes2009patchmatch, hays2007scene}. By nearest-neighbor searching and copying relevant patches, these methods can synthesize vivid textures for background inpainting tasks. However, since these methods cannot capture high-level semantics, it is hard for them to generate realistic structures for images with non-repetitive patterns (\eg faces).

With the advent of deep neural network techniques, some recent works~\cite{pathakCVPR16context,IizukaSIGGRAPH2017,yu2018free,yu2018generative,liu2018image} model the inpainting task as a conditional generation problem, which learns mapping functions between the input corrupted images and the ground truth images. These methods are able to learn meaningful semantics, so they can generate coherent structures for missing holes. However, since these methods do not effectively separate the structure and texture information, they often suffer from either over-smoothed boundaries or texture artifacts.

To solve this problem, some two-stage networks~\cite{yu2018generative,song2018spg,nazeri2019edgeconnect} are proposed. These methods recover missing structures in the first stage and generate the final results using the reconstructed information in the second stage. The method proposed in~\cite{yu2018generative} uses ground truth images as the labels of structure recovery. However, ground truth images contain high-frequency textures. These irrelevant details may mislead the structure reconstruction. Spg-net~\cite{song2018spg} predicts the semantic segmentation labels of the missing areas as structural information. However, regions with similar semantic labels may have different textures (e.g.\ the windows and walls of the same building), which creates difficulties for the final recovery. Using edge images as the structural guidance, EdgeConnect~\cite{nazeri2019edgeconnect} achieves good results even for some highly structured scenes. However, the distribution of edge images differs greatly from the distribution of the target images. In other words, the edge extractor discards too much useful information, such as image color, making it difficult to generate vivid textures.

In this paper, we propose a novel two-stage network StructureFlow for image inpainting. Our network consists of a structure reconstructor and a texture generator. To recover meaningful structures, we employ edge-preserved smooth images to represent the global structures of image scenes. Edge-preserved smooth methods~\cite{l0smoothing2011, xu2012structure} aim to remove high-frequency textures while retaining sharp edges and low-frequency structures. By using these images as the guidance of the structure reconstructor, the network is able to focus on recovering global structures without being disturbed by irrelevant texture information. After reconstructing the missing structures, the texture generator is used to synthesize high-frequency details. Since image neighborhoods with similar structures are highly correlated, the uncorrupted regions can be used to generate textures for missing regions. However, it is hard for convolutional neural networks to model long-term correlations~\cite{yu2018generative}. In order to establish a clear relationship between different regions, we propose to use appearance flow~\cite{zhou2016view} to sample features from regions with similar structures, as shown in Figure~\ref{fig:Formework}.
Since appearance flow is easily stuck within bad local minima in the inpainting task~\cite{yu2018generative}, in this work, we made two modifications to ensure the convergence of the training process. First, Gaussian sampling is employed instead of Bilinear sampling to expand the receptive field of the sampling operation. Second, we introduce a new loss function, called sampling correctness loss, to determine if the correct regions are sampled. 

Both subjective and objective experiments compared with several state-of-the-art methods show that our method can achieve competitive results. Furthermore, we perform ablation studies to verify our hypothesis and modifications. 
The main contributions of our paper can be summarized as:
\begin{itemize}
  \item We propose a structure reconstructor to generate edge-preserved smooth images as the global structure information.
  \item We introduce appearance flow to establish long-term corrections between missing regions and existing regions for vivid texture generation.
  \item To ease the optimization of appearance flow, we propose to use Gaussian sampling instead of Bilinear sampling and introduce a novel sampling correctness loss. 
  \item Experiments on multiple public datasets show that our method is able to achieve competitive results.
\end{itemize}

\section{Related Work}

\begin{figure*}[t]
\begin{center}
\includegraphics[width=0.9\linewidth]{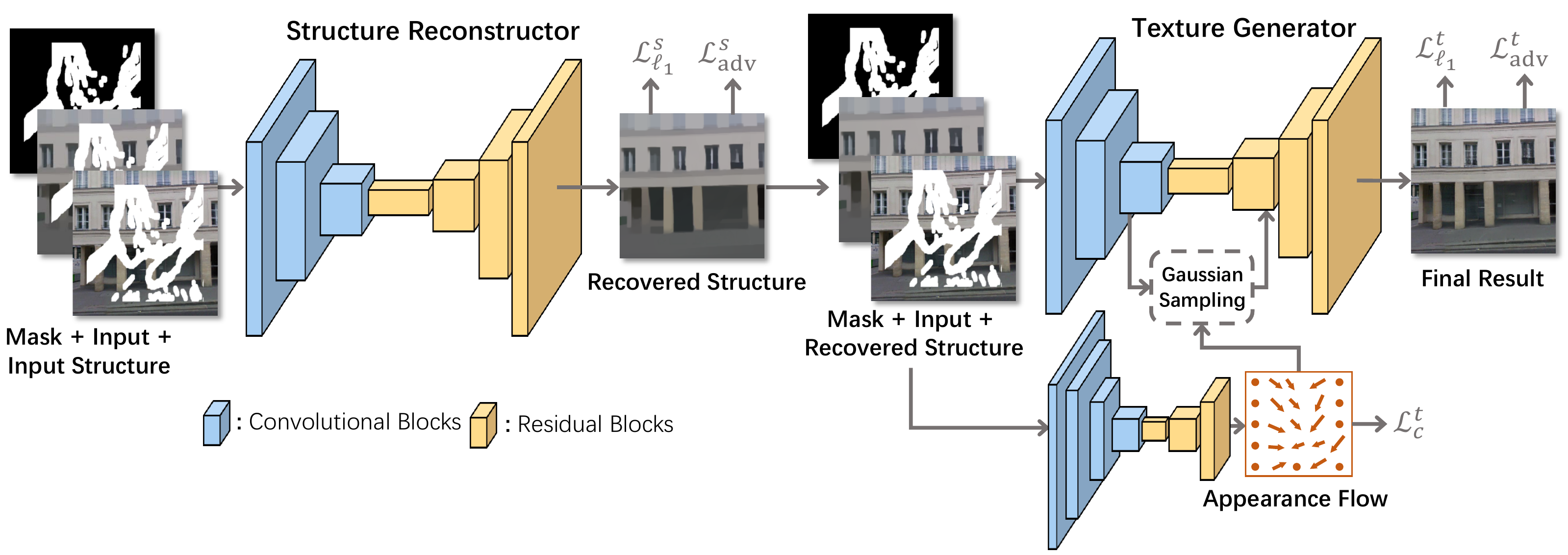}
\end{center}
   \caption{Overview of our StructureFlow. Our model first generates global structures (\ie edge-preserved smooth images) using structure reconstructor. Then texture generator is used to yield high-frequency details and output the final results. We add the appearance flow to our texture generator to sample features from existing regions.}
\label{fig:Network}
\end{figure*}

\subsection{Image Inpainting}
Existing image inpainting works can be roughly divided into two categories: methods using diffusion-based or patch-based techniques and methods using deep neural networks. Diffusion-based methods~\cite{bertalmio2000image, efros2001image} synthesize textures by propagating the neighborhood region appearance to the target holes. However, these methods can only deal with small holes in background inpainting tasks. They may fail to generate meaningful structures. Unlike the diffusion-based methods using only neighborhood pixels of missing holes, patch-based methods can take advantage of remote information to recover the lost areas. Patch-based methods~\cite{barnes2009patchmatch, hays2007scene,darabi2012image} fill target regions by searching and copying similar image patches from the uncorrupted regions of the source images. These methods can generate photo-realistic textures for relatively large missing holes. In order to find suitable image patches, bidirectional similarity~\cite{bidirectional} is proposed to capture more visual information and introduce less visual artifacts when calculating the patch similarity. To reduce the computational cost, PatchMatch~\cite{barnes2009patchmatch} designs a fast nearest neighbor searching algorithm using natural coherence in the imagery as prior information. However, these patch-based methods assume that the non-hole regions have similar semantic contents with the missing regions, which may not be true in some tasks such as face image inpainting. Therefore, they may work well in some images with repetitive structures but cannot generate reasonable results for images with unique structures.

Recently, many deep learning based methods have been proposed to model the inpainting task as a conditional generation problem. A significant advantage of these methods is that they are able to extract meaningful semantics from the corrupted images and generate new content for images.
Context Encoder~\cite{pathakCVPR16context} is one of the early attempts. It uses an encoder-decoder architecture to first extract features and then to reconstruct the outputs. However, this network struggles to maintain global consistency and often generate results with visual artifacts. Iizuka \etal~\cite{IizukaSIGGRAPH2017} solve this problem by using both local and global discriminators which are responsible for generating realistic alternative contents for missing holes and maintaining the coherency of competed images respectively. Yu \etal~\cite{yu2018generative} find that convolutional neural networks are ineffective in building long-term correlations. To solve this problem, they propose contextual attention to borrow features from remote regions. Liu \etal~\cite{liu2018image} believe the substituting pixels in the masked holes of the inputs introduce artifacts to the final results. Therefore, they propose partial convolutions to force the network to use valid pixels (uncorrupted pixels) only. Gated convolution~\cite{yu2018free} further generalizes this idea by extending the feature selecting mechanism to be learnable for each location across all layers. EdgeConnect proposed in paper~\cite{nazeri2019edgeconnect} has a similar motivation to our paper: generating reasonable structures by using additional prior information. EdgeConnect first recovers edge maps and then fills the missing regions in fine details. However, due to the limited representation ability of edge maps, this method may generate wrong details in the boundaries of objects.

\subsection{Optical Flow and Appearance Flow}
Optical flow is used to describe the motion of objects, surfaces, and edges between consecutive video frames. It has been widely used in video frame synthesis~\cite{liu2017voxelflow, wang2018vid2vid}, action recognition~\cite{simonyan2014two,wang2016temporal}, \etc. Optical flow estimation is an important task in computer vision. Many methods~\cite{horn1981determining, sun2008learning} have been proposed to accurately estimate optical flow between consecutive frames. Recently, some methods~\cite{dosovitskiy2015flownet, ilg2017flownet} solve this problem by training deep neural networks. However, these techniques require sufficient ground truth optical flow fields which are extremely difficult to obtain. Therefore, some synthetic optical flow datasets~\cite{dosovitskiy2015flownet} are created for training. Some other methods~\cite{liu2017video, meister2018unflow}  solve this problem by training the network in an unsupervised manner. However, many existing unsupervised optical flow estimation methods struggle to capture large motions. Some papers~\cite{liu2017video,ranjan2017optical} manage to use multi-scale approaches to improve the results. We believe it is due to the limited receptive field of Bilinear sampling. In this paper, we use Gaussian sampling as an improvement.

Appearance flow proposed by~\cite{zhou2016view} is used to generate target scenes (objects) from  source scenes (objects) using a flow-based method. It calculates the correlations between sources and targets to predict the 2-D coordinate vectors (\ie appearance flow fields). This idea can be used in image inpainting tasks. To generate realistic alternative contents for missing holes, one can reasonably "flow" pixels (features) from source regions to missing regions. In this paper, we improve the appearance flow in~\cite{zhou2016view} to make it suitable for image inpainting tasks.


\section{Our Approach}
The framework of our StructureFlow inpainting network is shown in Figure~\ref{fig:Network}. Our model consists of two parts: the structure reconstructor $G_s$ and the texture generator $G_t$. The structure reconstructor $G_s$ is used to predict missing structures, thereby generating the global structure image $\hat{\mathbf{S}}$. The texture generator $G_t$ draws details according to the reconstructed structures $\hat{\mathbf{S}}$ and outputs the final results $\hat{\mathbf{I}}$.

\subsection{Structure Reconstructor}
A major challenge of image inpainting tasks is to generate meaningful structures for missing regions. Therefore, we first design a structure reconstructor $G_s$ to recover global structures of the input images. The edge-preserved smooth methods~\cite{l0smoothing2011, xu2012structure} aim to remove high-frequency textures while retaining the sharp edges and low-frequency structures. Their results can well represent global structures. Let $\mathbf{I}_{gt}$ be the ground-truth image and $\mathbf{S}_{gt}$ be the edge-preserved smooth result of $\mathbf{I}_{gt}$. The processing of our structure reconstructor $G_s$ can be written as 
\begin{equation}
	\hat{\mathbf{S}}=G_{s}(\mathbf{I}_{in},\mathbf{S}_{in} ,\mathbf{M})
\end{equation}
where $\mathbf{M}$ is the mask of the input image $\mathbf{I}_{in}$. It is a binarized matrix where 1 represents the missing region and 0 represents the background. $\mathbf{S}_{in}=\mathbf{S}_{gt}\circ(\mathbf{1}-\mathbf{M})$ is the structures of $\mathbf{I}_{in}$. Here, $\circ$ denotes element-wise product. $\hat{\mathbf{S}}$ is the predicted structures.

The reconstruction loss of $G_s$ is defined as the $\ell_1$ distance between the predicted structures $\hat{\mathbf{S}}$ and the ground-truth structures $\mathbf{S}_{gt}$. 
\begin{equation}
	\mathcal{L}_{\ell_1}^s=\lVert \hat{\mathbf{S}} - \mathbf{S}_{gt} \rVert_1
\end{equation}
Meanwhile, to mimic the distributions of the target structures $\mathbf{S}_{gt}$, we  apply generative adversarial framework~\cite{goodfellow2014generative} to our structure reconstructor. The adversarial loss of $G_s$ can be written as 
\begin{align}
	\mathcal{L}_{adv}^s &= \mathbb{E}[\log(1-D_s(G_s(\mathbf{I}_{in}, \mathbf{S}_{in},\mathbf{M})))] \notag \\
	&+ \mathbb{E}[\log D_s(\mathbf{S}_{gt})]
	\label{eq:advr}	
\end{align}
where $D_s$ is the discriminator of the structure reconstructor. 
We jointly train the generator $G_s$ and discriminator $D_s$ using the following optimization.
\begin{equation}
	\min_{G_s}\max_{D_s}\mathcal{L}^s(G_s,D_s)=\lambda_{\ell_1}^s \mathcal{L}_{\ell_1}^s + \lambda_{adv}^s\mathcal{L}_{adv}^s
	\label{eq:total_recovery}
\end{equation}
where $\lambda_{\ell_1}^s$ and $\lambda_{adv}^s$ are regularization parameters. We set $\lambda_{\ell_1}^s=4$ and $\lambda_{adv}^s=1$ in all experiments.

\subsection{Texture Generator}
After obtaining the reconstructed structure image $\hat{\mathbf{S}}$, our texture generator $G_t$ is employed to yield vivid textures. The processing of the texture generator $G_t$ can be written as 
\begin{equation}
	\hat{\mathbf{I}} = G_t(\mathbf{I}_{in}, \hat{\mathbf{S}}, \mathbf{M})
\end{equation}
where $\hat{\mathbf{I}}$ denotes the final output result. 
We use $\ell_1$ loss to calculate the reconstruction error.
\begin{equation}
	\mathcal{L}_{\ell_1}^t =\lVert \hat{\mathbf{I}} - \mathbf{I}_{gt} \rVert_1
\end{equation}
To generate realistic results, we employ adversarial loss in our texture generator.
\begin{align}
	\mathcal{L}_{adv}^t &= \mathbb{E}[\log(1-D_t(G_t(\mathbf{I}_{in}, \hat{\mathbf{S}},\mathbf{M})))] \notag \\
	&+ \mathbb{E}[\log D_t(\mathbf{I}_{gt})]
	\label{eq:advr}	
\end{align}

\begin{table*}[]
\centering
\setlength\extrarowheight{2pt}
\resizebox{.95\textwidth}{!}{%
\begin{tabular}{c||c|c|c||c|c|c||c|c|c}
\hline
            & \multicolumn{3}{c||}{PSNR}                                                                              & \multicolumn{3}{c||}{SSIM}                                                                              & \multicolumn{3}{c}{FID}                                                                              \\ \hline
Mask        & 0-20\%                           & 20-40\%                          & 40-60\%                          & 0-20\%                           & 20-40\%                          & 40-60\%                          & 0-20\%                           & 20-40\%                          & 40-60\%                         \\ \hline
            &                                  &                                  &                                  &                                  &                                  &                                  &                                  &                                  &                                 \\ [-12pt] \hline 
CA          & 27.150                           & 20.001                           & 16.911                           & 0.9269                           & 0.7613                           & 0.5718                           & 4.8586                           & 18.4190                          & 37.9432                         \\ \hline
PConv       & 31.030                           & 23.673                           & 19.743                           & 0.9070                           & 0.7310                           & 0.5325                           & -                                & -                                & -                               \\ \hline
EdgeConnect & 29.972                           & 23.321                           & 19.641                           & 0.9603                           & 0.8600                           & 0.6916                           & 3.0097                           & 7.2635                           & \textbf{19.0003}                          \\ \hline
Ours        & \textbf{32.029} & \textbf{25.218} & \textbf{21.090} & \textbf{0.9738} & \textbf{0.9026} & \textbf{0.7561} & \textbf{2.9420} & \textbf{7.0354} & 22.3803 \\ \hline
\end{tabular}}
\caption{The evaluation results of CA~\cite{yu2018generative}, PConv~\cite{liu2018image}, EdgeConnect~\cite{nazeri2019edgeconnect}, and our model over dataset Places2~\cite{zhou2018places}.
Since the code and models of PConv are not available, we report the results presented in their paper.}
\label{Objective_Comparison}
\end{table*}

Since image regions with similar structures are highly related, it is possible to extract these correlations using the reconstructed structures $\hat{\mathbf{S}}$ for texture generation to improve the performance. However, convolutional neural networks are not effective for capturing long-term dependency~\cite{yu2018generative}. In order to establish a clear relationship between different regions, we introduce the appearance flow to our $G_t$.
As shown in Figure~\ref{fig:Network}, the appearance flow is used to warp the extracted features of the inputs. Thus, features containing vivid texture information can "flow" to the corrupted regions.

However, training the appearance flow in an unsupervised manner is a difficult task~\cite{liu2017video,ranjan2017optical}. The networks may struggle to capture large motions and stuck in a bad local minima. To tackle this problem, we first propose to use Gaussian sampling instead of Bilinear sampling to expand the receptive field. Then, we propose a sampling correctness loss to constraint the possible convergence results.

The sampling process calculates the gradients according to the input pixels (features). If the receptive field of the sampling operation is limited, only a few pixels can participate in the operation. Since the adjacent pixels (features) are often highly correlated, a large receptive field is required to obtain correct and stable gradients.
Therefore, Bilinear sampling with a very limited receptive field may not be suitable for tasks requiring establishing long-term correlations. To expand the receptive field, we use Gaussian sampling instead of Bilinear sampling in the appearance flow operation. The process of Gaussian sampling operation with kernel size $n$ can be written as 
\begin{equation}
\mathbf{F}_{o} = \sum_{i=1}^{n}\sum_{j=1}^{n}\frac{a_{i,j}}{\sum\nolimits_{i=1}^{n}\sum\nolimits_{j=1}^{n}a_{i,j}}\mathbf{F}_{i,j}
\label{gaussian}
\end{equation}
where $\mathbf{F}_{i,j}$ is the features around the sample center and $\mathbf{F}_o$ is the output feature. The weights $a_{i,j}$ is calculated as
\begin{equation}
	a_{i,j} =exp({-\frac{\Delta{h}^2 + \Delta{v}^2}{2\sigma^2}})
	\label{parameter}
\end{equation}
where $\Delta{h}$ and $\Delta{v}$ is the horizontal and vertical distance between the sampling center and feature $\mathbf{F}_{i,j}$ respectively. Parameter $\sigma$ is used to denote the variance of the Gaussian sampling kernel.

The proposed sampling correctness loss is used to constraint the appearance flow fields.
It determines whether the current sampled regions are "good" choices. 
We use the pre-trained VGG19 to calculate this loss. Specifically, we first calculate the VGG features of the input corrupted image $\mathbf{I}_{in}$ and the ground truth image $\mathbf{I}_{gt}$. Let $\mathbf{V}^{in}$ and $\mathbf{V}^{gt}$ be the features generated by a specific layer of VGG19. 
Symbol $M$ denotes a coordinate set containing the coordinates of missing areas, $N$ is the number of elements in set $M$.
Then, our sampling correctness loss calculate the relative cosine similarity between the ground truth features and the sampled features
\begin{equation}
	\mathcal{L}_{c}^{t} = \frac{1}{N}\sum_{(x,y)\in M}exp(-\frac{\mu(\mathbf{V}^{gt}_{x,y}, \mathbf{V}^{in}_{x+\Delta x, y+\Delta y})}{\mu_{x,y}^{max}})
\end{equation}
where $\mathbf{V}_{x+\Delta x, y+\Delta y}^{in}$ is the sampled feature calculated by our Gaussian sampling and $\mu(*)$ denotes the cosine similarity.
$\mu_{x,y}^{max}$ is a normalization term. For each feature $\mathbf{V}^{gt}_{x, y}$ where $(x,y)\in M$, we find the most similar feature from $\mathbf{V}_{in}$ and calculate their cosine similarity as $\mu_{x,y}^{max}$.
\begin{equation}
	\mu_{x,y}^{max} = \max\limits_{(x',y')\in \Omega}\mu(\mathbf{V}^{gt}_{x, y},\mathbf{V}^{in}_{x',y'})
\end{equation}
where $\Omega$ denotes a coordinate set containing all coordinates in $\mathbf{V}_{in}$.
Our texture generator is trained using the following optimization
\begin{equation}
	\min_{G_t}\max_{D_t}\mathcal{L}^t(G_t,D_t)=\lambda_{\ell_1}^t \mathcal{L}_{\ell_1}^t + \lambda_{c}^t\mathcal{L}_{c}^t + \lambda_{adv}^t\mathcal{L}_{adv}^t
	\label{eq:total_recovery}
\end{equation}
where $\lambda_{\ell_1}^t$, $\lambda_{c}^t$ and $\lambda_{adv}^t$ are the hyperparameters.
In our experiments, we set $\lambda_{\ell_1}^t=5$, $\lambda_{c}^t=0.25$ and $\lambda_{adv}^t=1$.

\section{Experiments}
\subsection{Implementation Details}
Basically, autoencoder structures are employed to design our generators $G_s$ and $G_t$.
Several residual blocks~\cite{he2016deep} are added to further process the features. 
For the appearance flow, we concatenate the warped features with the features obtained by convolutional blocks. The architecture of our discriminators is similar to that of BicycleGAN~\cite{zhu2017toward}. We use two PatchGANs~\cite{isola2017image} with different scales to predict real \vs fake for overlapping image patches with different sizes. In order to solve the notorious problem of instability training of generative adversarial networks, spectral normalization~\cite{miyato2018spectral} is used in our network. 

We train our model on three public datasets including Places2~\cite{zhou2018places}, Celeba~\cite{liu2015deep}, and Paris StreetView~\cite{doersch2012makes}. The most challenging dataset Places2 contains more than 10 million images comprising $400\texttt{+}$ unique scene categories. Celeba and Paris StreetView contain highly structured face and building images respectively. We use the irregular mask dataset provided by~\cite{liu2018image}. The mask images are classified based on their hole sizes relative to the entire image (\eg $0-20\%$ \etc).

We employ edge-preserved smooth method RTV~\cite{xu2012structure} to obtain the training labels of the structure reconstructor $G_t$. In RTV smooth method, parameter $\sigma$ is used to control the spatial scale of smooth windows, thereby controlling the maximum size of texture elements. In section~\ref{sec:ablation}, we explore the impact of $\sigma$ on the final results. We empirically find the best results obtained when we set $\sigma \approx 3$.

We train our model in stages.
First, the structure reconstructor $G_s$ and the texture generator $G_t$ are trained separately using the edge-preserved image $\mathbf{S}_{gt}$.
Then, we continue to fine-tune $G_t$ using the reconstructed structures $\mathbf{\hat{S}}$.
The network is trained using $256 \times 256$ images with batch size as 12.
We use the Adam optimizer~\cite{kingma2014adam} with learning rate as $10^{-4}$. 

\subsection{Comparisons}

\begin{figure*}[t]
\begin{center}
\includegraphics[width=1\linewidth]{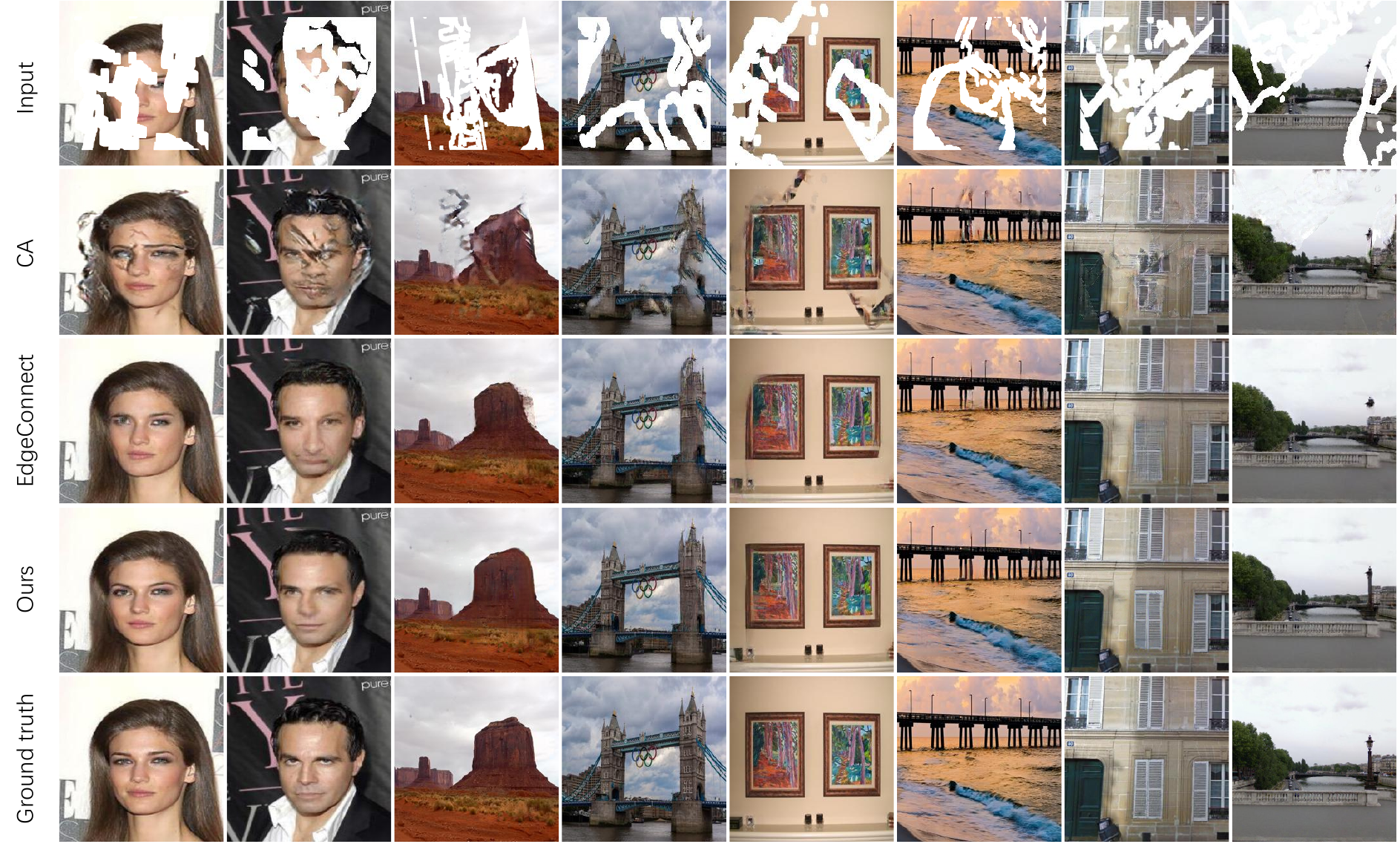}
\end{center}
   \caption{The qualitative comparisons with existing models. (From top to bottom) Input corrupted images, results of CA~\cite{yu2018generative}, results of EdgeConnect~\cite{nazeri2019edgeconnect}, results of our StructureFlow, and Ground truth images.}
 
\label{fig:compare}
\end{figure*}

\begin{figure}[t]
\begin{center}
\includegraphics[width=1\linewidth]{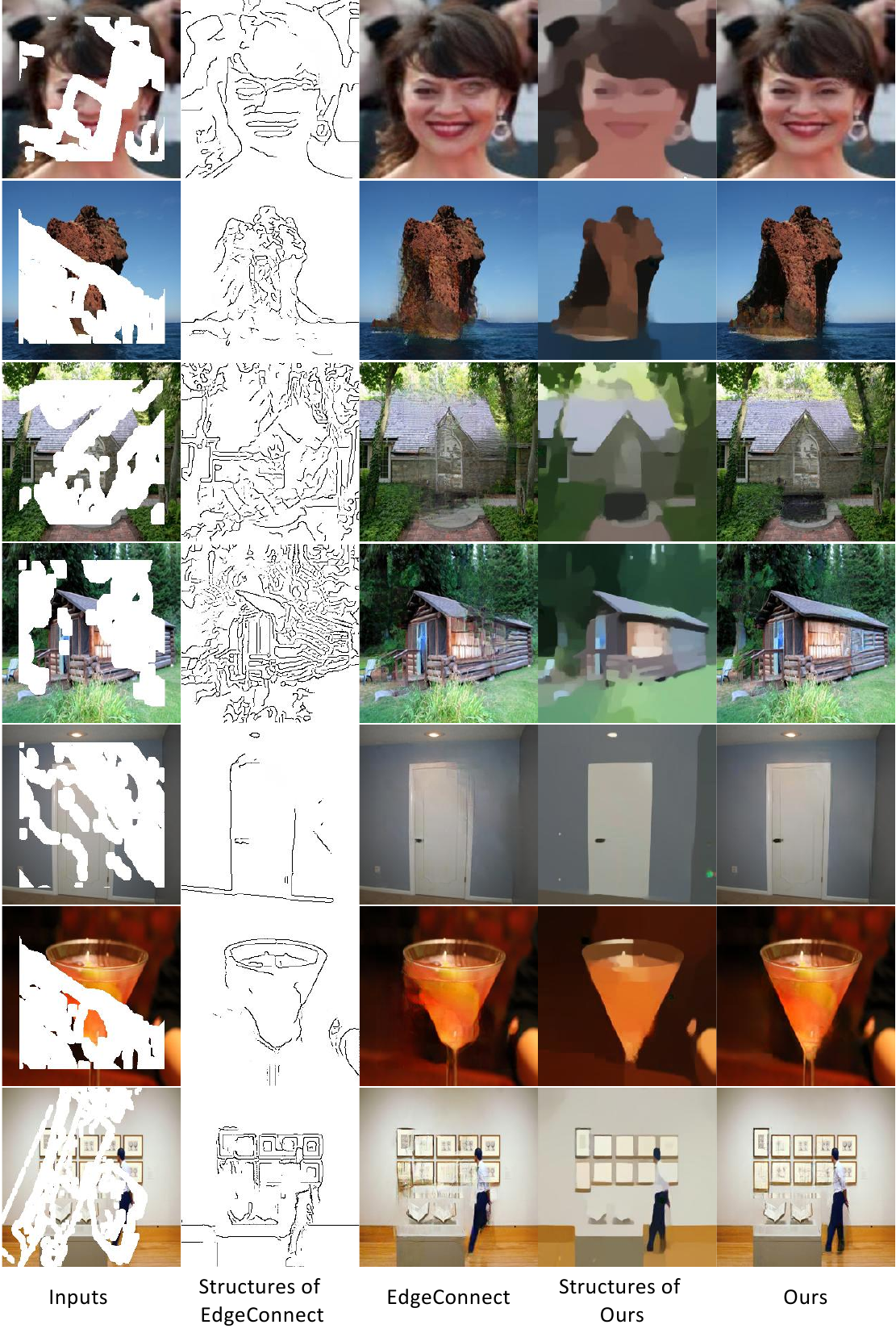}
\end{center}
\caption{The visual comparisons between EdgeConnect~\cite{nazeri2019edgeconnect} and our StructureFlow. 
(From left to right) Input corrupted images, reconstructed structures of EdgeConnect, inpainted results of EdgeConnect, reconstructed structures of our StructureFlow, inpainted results of our StructureFlow.}
\label{fig:ours}
\end{figure}
We subjectively and objectively compare our approach with several state-of-the-art methods including Contextual Attention (CA)~\cite{yu2018generative}, Partial Convolution (PConv)~\cite{liu2018image} and EdgeConnect~\cite{nazeri2019edgeconnect}.

\textbf{Objective comparisons}
Image inpainting tasks lack specialized quantitative evaluation metrics. In order to compare the results as accurately as possible, we employ two types of metrics: distortion measurement metrics and perceptual quality measurement metrics. Structural similarity index (SSIM) and peak signal-to-noise ratio (PSNR) assume that the ideal recovered results are exactly the same as the target images. They are used to measure the distortions of the results. Fr\'echet Inception Distance (FID)~\cite{heusel2017gans} calculates the Wasserstein-2 distance between two distributions. Therefore, it can indicate the perceptual quality of the results. In this paper, we use the pre-trained Inception-V3 model to extract features of real and inpainted images when calculating FID scores. The final evaluation results over Places2 are reported in Table~\ref{Objective_Comparison}. 
We calculate the statistics over $10k$ random images in the test set.
It can be seen that our model achieves competitive results compared with other models. 
 
\textbf{Subjective comparisons}
We implement a human subjective study on the Amazon Mechanical Turk (MTurk). We ask volunteers to choose the more realistic image from image pairs of real and generated images.
For each dataset, we randomly select $600$ images and assign them random mask ratios from $0\%-60\%$ for the evaluation.
Each image is compared $5$ times by different volunteers.
The evaluation results are shown in Table~\ref{tab:sub}. 
Our model achieves better results than the competitors in the highly-structured scenes,
such as face dataset Celeba and street view dataset Paris.
This indicates that our model can generate meaningful structures for missing regions.
We also achieve competitive results in dataset Places2.

\begin{table}[]
\setlength\extrarowheight{2pt}
\centering
\begin{tabular}{c||c|c|c}
\hline
       & CA      & EdgeConnect    & Ours    \\ \hline
       &         &         &         \\[-12pt] \hline  
Celeba & 5.68\%  & 26.28\% & \textbf{32.04\%} \\ \hline
Paris  & 17.36\% & 33.44\% & \textbf{33.68\%} \\ \hline
Places2 & 8.72\%  & \textbf{26.36\%} & 23.56\% \\ \hline
\end{tabular}
\caption{The evaluation results of user study. The volunteers are asked to select the more realistic image from image pairs of real and generated images. The fooling rate is provided in the table.}
\label{tab:sub}
\end{table}

Figure~\ref{fig:compare} shows some example results of different models. It can be seen that the results of CA suffer from artifacts, which means that this method may struggle to balance the generation of textures and structures. EdgeConnect is able to recover correct global structures. However, it may generate wrong details at the edges of objects. Our method can generate meaningful structures as well as vivid textures. We also provide the reconstructed structures of EdgeConnect and our model in Figure~\ref{fig:ours}. We find that the edge maps loss too much useful information, such as image color when recovering the global structures. Therefore, EdgeConnect may fill incorrect details for some missing areas. Meanwhile, edges of different objects may be mixed together in edge maps, which makes it difficult to generate textures. In contrast, our edge-preserved smooth images can well represent the structures of images. Therefore, our model can well balance structure reconstruction and texture generation. Photo-realistic results are obtained even for some highly structured images with large hole ratios.

\begin{figure}[t]
	\begin{center}
		\includegraphics[width=1\linewidth]{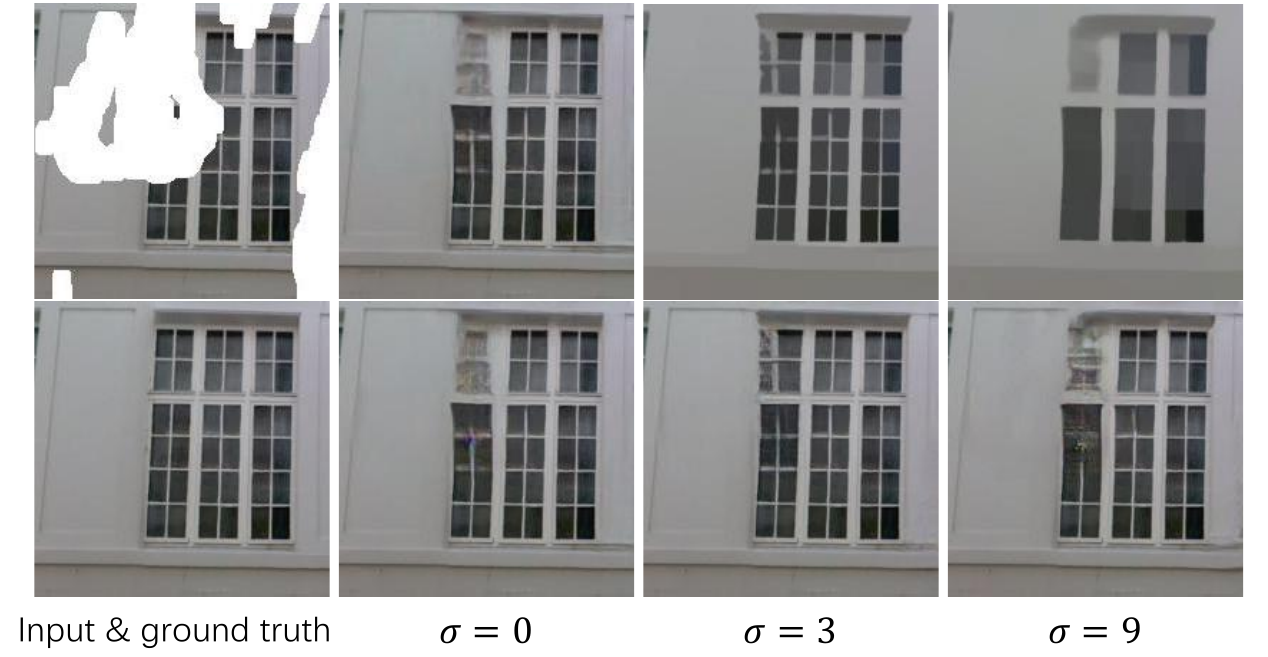}
	\end{center}
	\caption{The influence of the parameter $\sigma$ in RTV edge-preserved smooth method on final results. The last three columns show results of models trained by smooth images generated with $\sigma=0,3,9$, where the first row shows the results of structure reconstructor and the second row shows the generated results. }
	\label{fig:multi-smooth}
\end{figure}

\begin{table}[]
\centering
    \setlength\extrarowheight{1pt}
    \resizebox{.42\textwidth}{!}{%
        \begin{tabular}{cc||p{45pt}<{\centering}|p{45pt}<{\centering}}
            \hline
            &               & PSNR & SSIM \\ \cline{3-4} 
             \hline
            & & &\\[-11pt] \hline 
            \multicolumn{1}{c|}{\multirow{4}{*}{\rotatebox{90}{Paris}}}  & w/o Structure & 28.46    & 0.8879      \\ \cline{2-4}
            \multicolumn{1}{c|}{}                        & w/o Smooth    & 28.41       & 0.8848      \\ \cline{2-4}
            \multicolumn{1}{c|}{}                        & w/o Flow      & 28.77       & 0.8906      \\ \cline{2-4}
            \multicolumn{1}{c|}{}                        & StructureFlow & \textbf{29.25}       & \textbf{0.8979}   \\ \hline
            \multicolumn{1}{c|}{}  & & &\\[-11pt] \hline 
            \multicolumn{1}{c|}{\multirow{4}{*}{\rotatebox{90}{Celeba}}} & w/o Structure & 29.42     & 0.9324      \\ \cline{2-4}
            \multicolumn{1}{c|}{}                        & w/o Smooth    & 29.61       & 0.9335      \\ \cline{2-4}
            \multicolumn{1}{c|}{}                        & w/o Flow      & 29.91       & 0.9368      \\ \cline{2-4}
            \multicolumn{1}{c|}{}                        & StructureFlow & \textbf{30.31}       & \textbf{0.9420}      \\ \hline
        \end{tabular}%
    }
    \caption{The evaluation results of ablation studies. We provide the statistics of four models: the model trained without the structure reconstructor (\ie w/o Structure), the model trained using ground truth images as the labels of the structure reconstructor (\ie w/o Smooth), the model trained without the appearance flow operation (\ie w/o Flow) and our full model (\ie StructureFlow). The statistics are based on random masks with mask ratio $0\%$-$60\%$.}
    \label{ablation}
\end{table}

\subsection{Ablation Studies}\label{sec:ablation}
In this section, we analyze how each component of our StructureFlow contributes to the final performance from two perspectives: structures and appearance flow.

\textbf{Structure Ablation}
In this paper, we assume that the structure information is important for image inpainting tasks.
Therefore, we first reconstruct structures and use them as prior information to generate the final results.
To verify this assumption, we remove our structure reconstructor and train an inpainting model with only the texture generator.
The corrupted images along with its masks are directly inputted into the model. 
Please note that we also keep appearance flow in the network for fair comparisons.
The results are shown in Table~\ref{ablation}. 
It can be seen that our structure reconstructor can bring stable performance gain to the model.

\begin{table}[t]
	\setlength\extrarowheight{2pt}
	\resizebox{.48\textwidth}{!}{%
		\centering
		\begin{tabular}{c||c|c|c|c|c}
			\hline
			& $\sigma=0$      &$\sigma=1$ & $\sigma=3$  & $\sigma=6$ & $\sigma=9$      \\ \hline
			&        &        &        &        &        \\[-12pt]  \hline
			PSNR & 28.41  & 28.81  & \textbf{29.25}  & 29.14  & 28.98  \\ \hline
			SSIM & 0.8848 & 0.8896 & \textbf{0.8979} & 0.8962 & 0.8990 \\ \hline
	\end{tabular}}
	\caption{The evaluation results over dataset Paris of models trained using edge-preserved images generated with $\sigma=0,1,3,6,9$. The statistics are based on random masks with mask ratio $0\%$-$60\%$.}
	\label{smooth}
\end{table}

Then we turn our attention to the edge-preserved smooth images. We believe the edge-preserved smooth images are able to represent the structures since the smooth operations remove high-frequency textures. To verify this, we train a model using ground truth images $\mathbf{I}_{gt}$ as the labels of the structure reconstructor. The results can be found in Table~\ref{ablation}. Compared with StructureFlow, we can find that using images containing high-frequency textures as structures leads to performance degradation.

However, it is difficult to accurately distinguish the textures and the structures of an image. What is the appropriate degree of smooth operation? We find there exists a trade-off between the structure reconstructor and the texture generator. If very few textures are removed, the structure reconstruction will be more difficult, since it needs to recover more information. However, the texture generation will be easier. Therefore, we need to balance the difficulties of these two tasks to achieve better results. We use $\sigma$ in RTV~\cite{xu2012structure} smooth method to control the maximum size of texture elements in $\mathbf{S}_{gt}$. Smoother results are obtained with larger $\sigma$ value. We train our StructureFlow using smooth images generated from $\sigma=0,1,3,6,9$. The evaluation results over dataset Paris are shown in Table~\ref{smooth}.  It can be seen that the best results are obtained when $\sigma=3$. Both too small and too large $\sigma$ values lead to model performance degradation. An example can be found in Figure~\ref{fig:multi-smooth}. When $\sigma=0$, the structure reconstructor fail to generate reasonable structures, as it is disturbed by irrelevant texture information. The texture generator fails to yield realistic images when trained with $\sigma=9$ since some useful structural information is removed.

\begin{figure}[t]
	\begin{center}
		\includegraphics[width=1\linewidth]{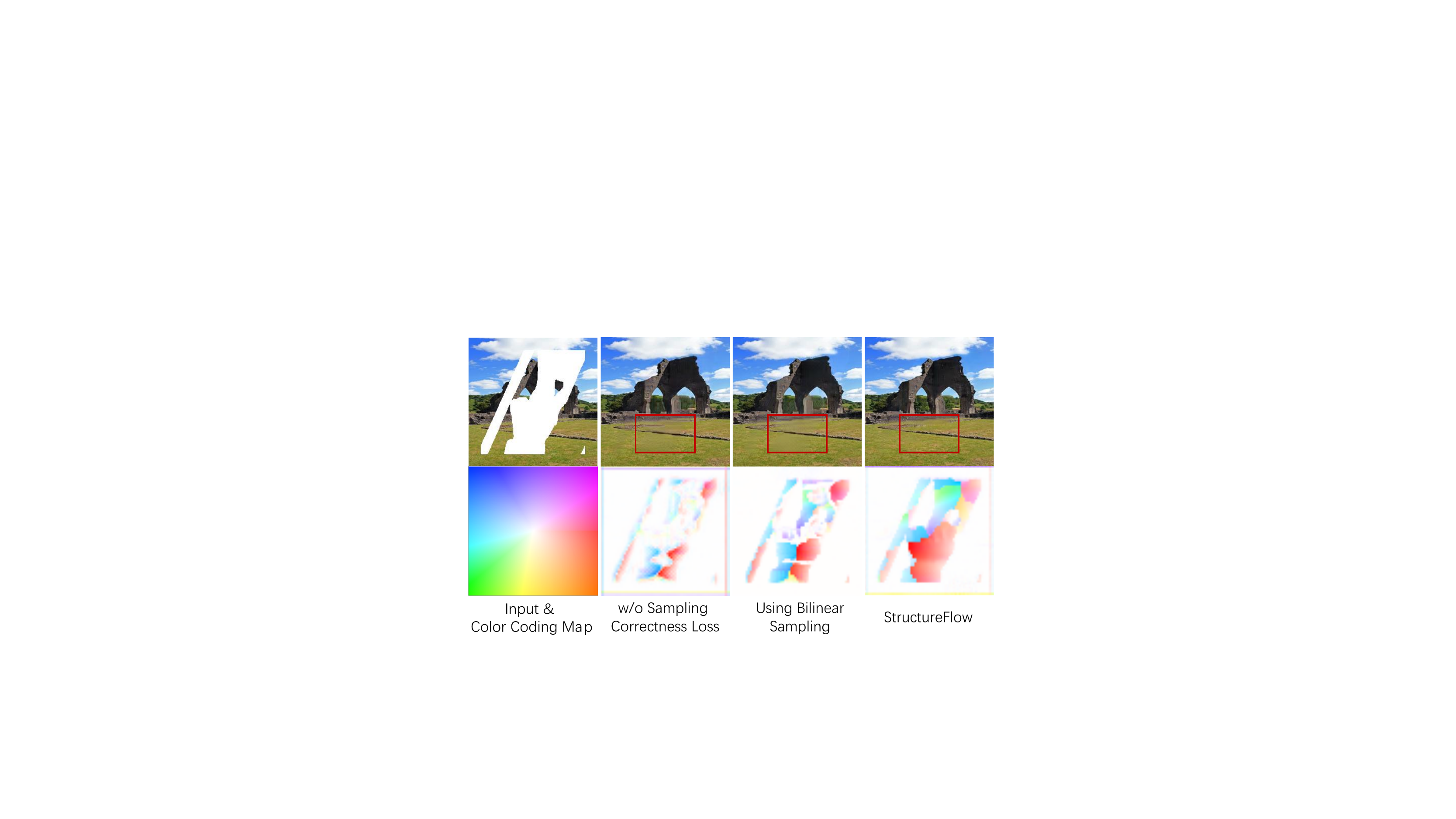}
	\end{center}
	\caption{Ablation studies of Gaussian sampling and the sampling correctness loss. The appearance flow fields are visualized using the provided color coding map. Flow direction is encoded with color and magnitude with color intensity.}
	\label{fig:ablation_flow}
\end{figure}

\textbf{Flow Ablation}
In this ablation study, we first evaluate the performance gain bought by our appearance flow. Then, we illustrate the effectiveness of Gaussian sampling and the sampling correctness loss.

To verify the validity of our appearance flow,
we train a model without using the appearance flow blocks in the texture generator. 
The evaluation results can be found in Table~\ref{ablation}.
It can be seen that our StructureFlow has better performance than the model trained without the appearance flow operation, which means that our appearance flow can help with the texture generation and improve model performance.

Next, we test our Gaussian sampling and the sampling correctness loss. Two models are trained for this ablation study: a model trained using Bilinear sampling in the warp operation of appearance flow and a model trained without using the sampling correctness loss. Figure~\ref{fig:ablation_flow} shows the appearance flow fields obtained by these models. It can be seen that the model trained without using the sampling correctness loss is unable to sample correct features for large missing regions. 
Bilinear sampling also fails to capture long-term correlations. Our StructureFlow obtains a reasonable flow field and generates realistic textures for missing regions. 

\subsection{User case}
\begin{figure}[t]
	\begin{center}
		\includegraphics[width=1\linewidth]{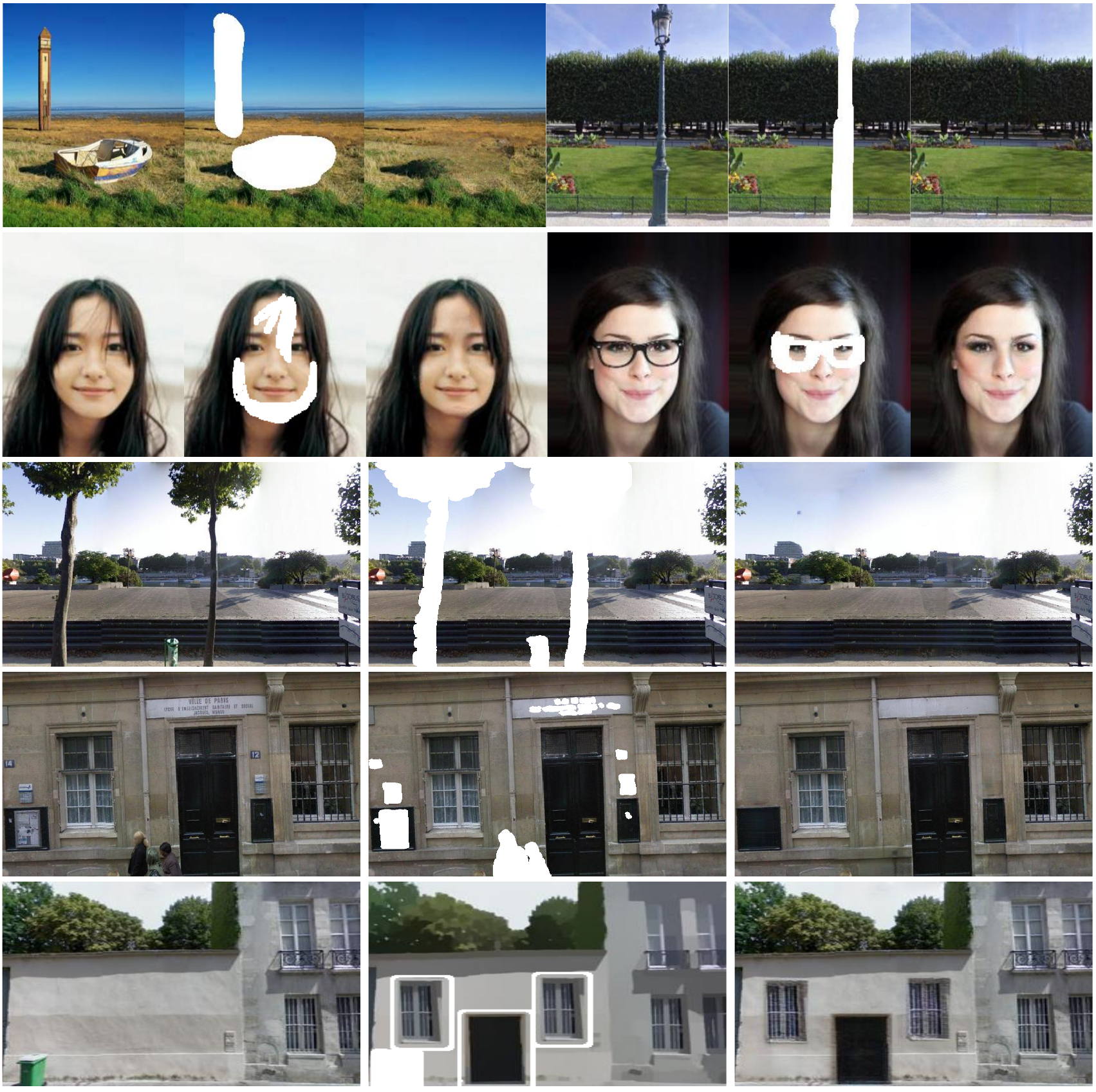}
	\end{center}
	\caption{Examples of object removing and image editing using our StructureFlow. Our model is able to generate realistic alternative contents for the missing regions.}
	\label{fig:usage}
\end{figure}

Our method can be used for some image editing applications. Figure~\ref{fig:usage} provides some usage examples. Users can remove the unwanted objects by interactively drawing masks in the inputs. Our model is able to generate realistic alternative contents for the missing regions. 
In addition, by directly editing the structure images, users can copy or add new objects and contents to images.

\section{Conclusion}
In this paper,
we propose an effective structure-aware framework for recovering corrupted images with meaningful structures and vivid textures.
Our method divides the inpainting task into two subtasks: structure reconstruction and texture generation.
We demonstrate that edge-preserved smooth images can well represent the global structure information and play an important role in inpainting tasks.
As for texture generation,
we use appearance flow to sample features from relative regions.
We verify that our flow operation can bring stable performance gain to the final results.
Our method can obtain competitive results compared with several state-of-the-art methods. Our source code is available at: \url{https://github.com/RenYurui/StructureFlow}.


\bigbreak

\noindent
\textbf{Acknowledgements.} 
This work was supported by National Engineering Laboratory for Video Technology-Shenzhen Division, Shenzhen Municipal Science and Technology Program (JCYJ20170818141146428), and Shenzhen Key Laboratory for Intelligent Multimedia and Virtual Reality (ZDSYS201703031405467). In addition, we thank the anonymous reviewers for their valuable comments.

{\small
\bibliographystyle{ieee_fullname.bst}
\bibliography{egbib.bib}
}

\end{document}